\pdfoutput=1
\documentclass[11pt]{article}

\usepackage{acl}

\usepackage{times}
\usepackage{latexsym}

\usepackage[T1]{fontenc}
\usepackage[utf8]{inputenc}

\usepackage{microtype}

\usepackage{multirow}
\usepackage{hyperref}
\usepackage{tabularx}
\usepackage{booktabs}
\usepackage{graphicx}
\graphicspath{{./images/}}
\usepackage{amssymb}
\usepackage{enumitem}
\usepackage{makecell}
\usepackage[section]{placeins}

\newcommand{\bto}[1]{}

\title{Corpus-Guided Contrast Sets for Morphosyntactic Feature Detection in Low-Resource English Varieties}

\author{Tessa Masis \\
  \emph{they/them/theirs} \And
  Anissa Neal \\
  \emph{she/her/hers} \And
  Lisa Green \\
  \emph{she/her/hers} \And
  Brendan O'Connor \\
  \emph{he/him/his} \AND
  {\normalfont University of Massachusetts Amherst} \\
  \texttt{\{tmasis,brenocon\}@cs.umass.edu} \\
  \texttt{\{anneal,lgreen\}@linguist.umass.edu}
  }

\begin{document}
\maketitle
\begin{abstract}
The study of language variation examines how language varies between and within different groups of speakers, shedding light on how we use language to construct identities and how social contexts affect language use. A common method is to identify instances of a certain linguistic feature---say, the zero copula construction---in a corpus, and analyze the feature's distribution across speakers, topics, and other variables, to either gain a qualitative understanding of the feature's function or systematically measure variation. 
%For example, identifying and analyzing occurrences of the feature zero copula, defined as the omission of a copula i.e. "She $\varnothing$ good". 
In this paper, we explore the challenging task of automatic morphosyntactic feature detection in low-resource English varieties.
We present a human-in-the-loop approach to generate and filter effective contrast sets via corpus-guided edits. We show that our approach improves feature detection for both Indian English and African American English, demonstrate how it can assist linguistic research, and release our fine-tuned models for use by other researchers.

\end{abstract}

\section{Introduction}
\begin{figure}[t]
\centering
\fbox{\includegraphics[width=.45\textwidth]{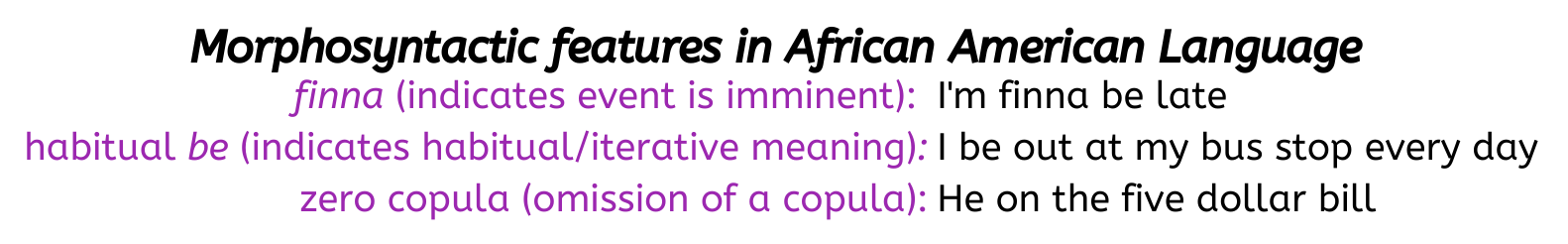}}\\
\vspace{.1cm}
\fbox{\includegraphics[width=.45\textwidth]{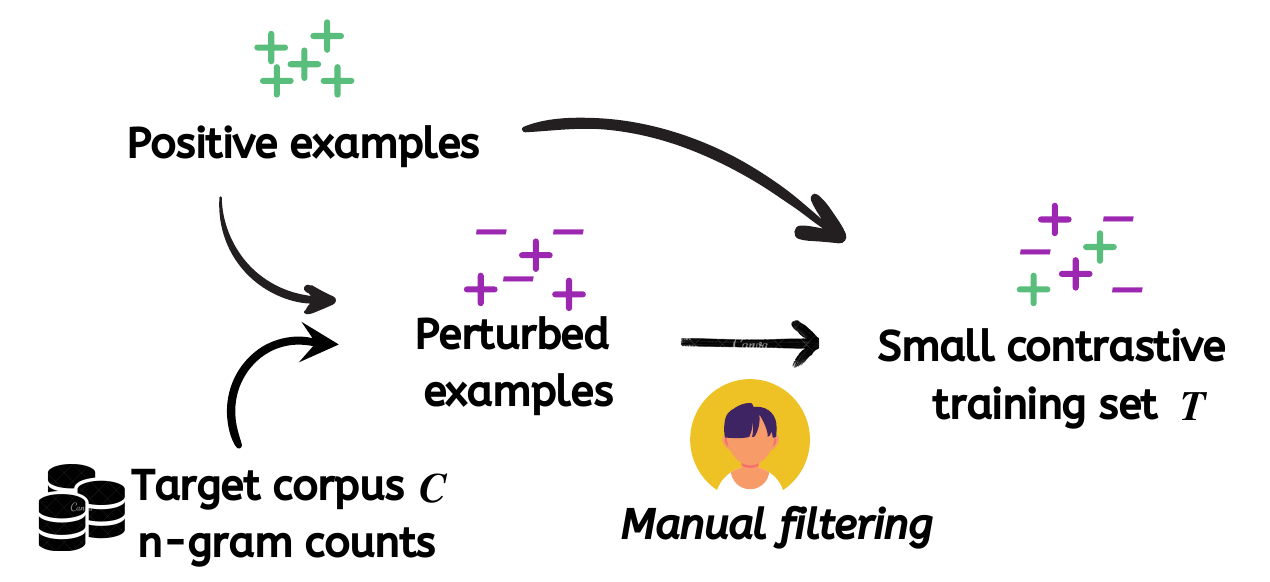}}
\caption{Top: Example features. Bottom: Our approach to generate contrast sets for feature detection.}
\label{fig:diagram1}
\end{figure}

%\bto{i attempted to define `feature'. everything needs to be comprehensible to someone who has never heard of variationist socioling or `feature' as in linguistics, not ML.}
Linguistic \emph{features}---such as specific phonological, syntactic, or lexical phenomena that may be associated with a language variety---are widely used by sociolinguists to quantify linguistic variation between speakers through feature frequency measurements \citep{renn09, grieser19, craig06}, even if subject to certain limitations \cite{green17}.
Since manual annotation is limited due to the required expert human labor, automatic methods are a valuable alternative \citep{grieve10, jones15, eisenstein15, nguyen16}. 
However, accurately detecting morphosyntactic features (e.g.\ Figure \ref{fig:diagram1})
remains an open challenge, especially in informal genres such as transcripts and social media, and in low resource nonstandard languages.
We explore fine-tuning pretrained language models (LMs) for utterance-level classification of a feature by training on a \emph{contrast set}---a small collection of positive and negative examples that are highly similar---as recently introduced by \citet{demszky21}. 
%\bto{`contrast set' was never defined. is my defn correct? what do we want to communicate with the term?}

Our work makes the following contributions:
\setlist{nosep}
\begin{itemize}[noitemsep]
    \item We propose a method for generating morphosyntactically contrastive training data, combining corpus-driven edits and human-in-the-loop filtering
    (\hyperlink{section.4}{\S4}). 
    \item We evaluate our method's ability to detect features against new baselines on three datasets, encompassing two Englishes (Indian English (IndE) and African American English (AAE)) and two centuries of speakers, and show that our best method outperforms prior work by up to 16 points in Prec@100 scores (\hyperlink{section.5}{\S5}).
    \item For further validation, we confirm and extend the findings of sociolinguistic studies of AAE which use manual feature annotation to examine if feature use aligns with social factors like age and gender (\hyperlink{section.6}{\S6}).
    \item Finally, we release training data and models for detecting 10 features in IndE and 17 in AAE.\footnote{\texttt{https://github.com/slanglab/CGEdit}}
\end{itemize}

\section{Related Work}
\textbf{Feature detection.}
Detecting morphosyntactic features in low-resource domains presents significant challenges. Rule-based approaches have used sequences of unigrams and POS tags to identify syntactic features \citep{blodgett16}, but many features cannot be defined by sequences and the tags may be unreliable. More recently, machine learning has been used for feature detection by training domain-specific LMs with synthetically augmented data \citep{santiago22}, fine-tuning pretrained LMs with contrast sets \citep{demszky21}, or manually filtering results from noisy classifiers \citep{austen17}. While prior work has only considered one language variety at a time and primarily evaluated with labeled test sets, we examine performance on multiple language varieties and analyze external sociolinguistic validity.

\textbf{Contrast set generation.} 
Manual generation of contrast sets has mostly been used for semantic tasks \citep{staliunaite17, mahler17, gardner20}, and occasionally for morphosyntactic tasks \citep{demszky21}. Unlike these approaches, our proposed method generates a \emph{morphosyntactically diverse} contrast set via a corpus-guided edit system. Data augmentation methods for automatic generation of contrast sets include random edits \citep{smith05, alleman21}, which cannot target specific linguistic features, or informed edits \citep{burlot17, sennrich17, gulordava18, miao20, ross21}, which require syntactic or semantic annotations that are not easily available for datasets with nonstandard languages.

\section{Task and Data}
\subsection{Morphosyntactic feature detection}
Given a training set $T$, target corpus $C$, and morphosyntactic features $F$,
for each $f \in F$ we model
\begin{equation}
P(f_x = 1 \mid T, x),
\end{equation}
where $f_x \in \{0,1\}$ indicates the utterance $x \in C$ contains the feature when $f_x=1$. An utterance may contain multiple features. 

\subsection{Language Varieties and Data}
We consider two English varieties, IndE and AAE, each with their own target corpora $C$ and feature inventories $F$; see Appendix \ref{sec:app1} for feature lists. 

\textbf{Indian English.}
The International Corpus of English (\textsc{ice}) \citep{greenbaum96} is a collection of national and regional English varieties, and contains IndE material produced after 1989. The \textsc{ice}-India subcorpus that our study uses is the complete subset of spontaneous spoken dialogues (21,759 utterances). We use manual annotations of 10 syntactic features from \citet{lange12}.

\textbf{African American English.}
We use two unlabeled AAE corpora. The first is the Corpus of Regional African American Language (\textsc{coraal}) \citep{kendall21}, which contains sociolinguistic interviews with AAE speakers from 1968-2017 from six US sites (152,069 utterances). The second is Born in Slavery: Slave Narratives from the Federal Writers' Project, 1936-38 (\textsc{fwp}) \citep{fwp}, a digital archive containing over 2,300 ex-slave narratives, with speakers from 17 US states (148,018 utterances).\footnote{Given authenticity and reliability concerns about \textsc{fwp} \citep{maynor88, wolfram90}, we primarily use it to evaluate our method, and not to pursue linguistic questions about Early African American English.}

We examine 17 AAE features, sourced from \citet{green02} and \citet{koenecke20}; examples of three features are in Figure \ref{fig:diagram1}, and a complete list is in Appendix \ref{sec:app1}. During evaluation, we manually annotated the top 100 utterances per AAE feature, for each corpus, for the Prec@100 scores in \hyperlink{section.5}{\S5}.

\section{\textsc{CGEdit}: Corpus-Guided Edits}
\subsection{Motivation}
Our method starts with a seed set of positive examples illustrating a feature, then uses corpus $n$-gram statistics to generate proposed negative (and additional positive) examples, which require manual filtering by the user to define the final training set. A major motivation is speed and ease of use---it is easier to filter candidate examples than to manually write all the examples, as in \citet{demszky21}.

At the same time, we believe negative examples should be intelligently synthesized. A morphosyntactic feature is beholden to its syntactic constraints (i.e.\ word order, co-occurrence requirements); if a sentence does not follow these constraints then it is not an instance of the feature \citep[Ch. 5.2]{wilson11}. For example, an instance of zero copula must have a noun phrase immediately followed by a predicate and must not have a copula. The positive example in Figure \ref{fig:diagram2} obeys these syntactic constraints while the negatives do not. Unlike previous work which uses constraints to detect or generate positive instances, \emph{we generate negative examples which minimally violate these constraints} to create a contrast set that defines a tight decision boundary. Based on the view that good syntax is largely independent from meaning \citep{chomsky57}, we argue that focusing on syntactic constraint violation is a useful first step. While potentially valuable, semantic-preserving edits are beyond the scope of this work.

% \bto{what advantage does this give? why do we not attempt to prserve sem meaning? i feel like this is missing the core reason for our approach... it's something like, you dont have to think as much in our approach bcs reviewing candidates is a lot easier then generating examples from scratch.  the syn vs sem argument is relevant only as a rebuttal to someone who says, but you need to preserve sem similarity if you want good contrastive examples.}

\subsection{Method}
\textbf{Training data.}
We briefly describe how the contrast sets are generated (Figure \ref{fig:diagram2}; see Appendix \ref{sec:app2} for details).
For a single feature, the input is a small set $P$ of 5 positive examples constructed by the authors
and an unlabeled target corpus $C$ to compute n-gram statistics. The output is a contrast set $T$ consisting of both $P$ plus semi-synthetic positive and negative examples. \bto{changed}

The first step proposes candidate examples by perturbing words in positive examples through corpus-guided local edits. For each overlapping 3-gram $t$ in a positive example $p$,
we perturb it by swapping $t$ for a new 2-, 3-, or 4-gram $t'$ that is both similar to $t$, and has a high frequency in target corpus $C$.
Similarity is defined as having 0 to 1 subtoken difference between $t$ and $t'$.\footnote{Specifically, the set difference between subtoken sets $set(t)$ and $set(t')$ must have cardinality 0 or 1; thus a 2-gram $t'$ represents a (sub)token deletion, a 4-gram an insertion, and perturbations may change order as well. Since only a single 3-gram is changed, the resulting perturbed utterance has a low edit distance to the original.}
This step typically produces 10-50 perturbed examples, which may or may not have the feature. Our corpus-guided edits are effective because they generate plausible sentences with targeted edits, while  random edits often propose ungrammatical output.\footnote{While our $n$-gram swapping heuristic is straightforward, generating from a $C$-specific language model could be an interesting alternative in future work.}

In the second step, the perturbed examples are manually filtered so that only 2 positive and 3 negative examples are retained for each original $p$. Both $p$ and the new examples are included in the final training set $T$. This step takes 30-60 seconds per $p$, and was performed by the first author.  \bto{last clause follows what i wrote 2 paragraphs above}
%\bto{need to say how many original positives per feature and how many proposals per orig positive---even if only a rough range like `10-20` or `50-150` (right now i have no idea)}\bto{they're gonna ask---what if you took 30-90 seconds to generate ManualGen negatives? can we answer that?}

%We refer to our approach as \textsc{CGEdit}.

\textbf{Models.}
We fine-tune multiheaded BERT models, where each head is a binary classifier for a single feature \citep{devlin19}. We use two sets of models in our experiments, where a set shares a language variety, a feature inventory $F$, target corpora $C$ (i.e. test set for our results in Table 1), and a BERT variant (\emph{bert-base-uncased} for IndE, \emph{bert-base-cased} for AAE, selected based on preliminary experiments). The only variation between models \emph{within} a set is the approach used to generate the training set $T$. Models were fine-tuned with cross-entropy loss for 500 epochs using the Adam optimizer, batch size of 64, and learning rate of $10^{-5}$, warmed up over the first 150 epochs.\footnote{Early experiments indicated that class-balanced loss did not improve scores.}

\begin{figure}[t]
\centering
\fbox{\includegraphics[width=.36\textwidth]{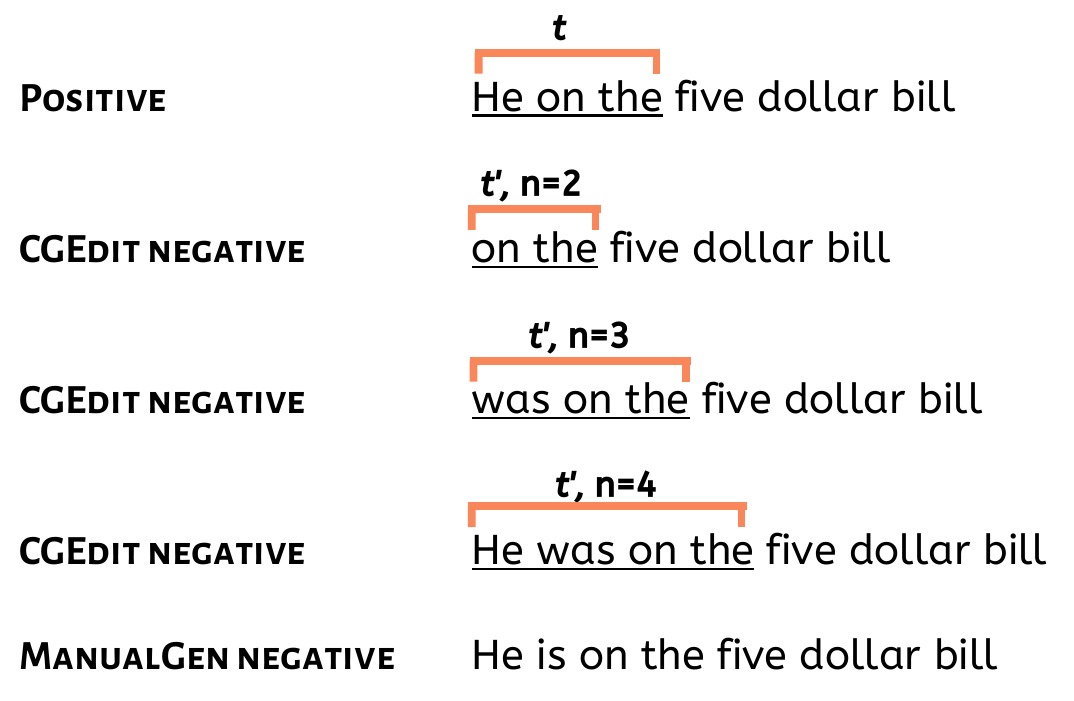}}
\caption{Examples of negative examples generated via our approach, compared to a semantically-matched, manually created example (\textsc{ManualGen}).}
\label{fig:diagram2}
\end{figure}

\section{Results and Analysis}
\textbf{Baselines.} We compare our approach (which we refer to as \textsc{CGEdit}) to several baseline methods, 
\bto{NEW, is this right?:}
all of which take the same seed set of positive examples $P$ then add negative examples to complete the training set. Examples in $P$ were sourced from \citet{demszky21} for IndE and crafted by the authors for AAE.

\textsc{ManualGen:} The approach used in \citet{demszky21}. This method involves manually generating negatives by modifying positive examples so they are (1) semantically-similar Mainstream American English versions, and (2) do not have the feature (see Figure \ref{fig:diagram2}); see discussion in \hyperlink{section.4}{\S4.1}. Next, we also test two methods to completely automatically generate negative examples:

\textsc{AutoGen:} This approach automatically generates negative examples by dividing a positive example $p$ into n-grams and shuffling the n-grams. 

\textsc{AutoID:} Automatic identification randomly chooses unlabeled examples from target corpus $C$ as the negatives. The assumption that unlabeled examples are negatives with class label noise underpins contrastive learning \citep{chen20} and PU learning methods \citep{bekker20}.

\bto{where do we get $P$s from?  for AAE we make them up ourselves, sure---we should say so. but for IndE didn'y they come from the Demszky paper?  was it always $|P|=5$, as implied by Sec 4 now?}

\begin{table*}
\centering
\begin{tabularx}{.95\textwidth}{l c c c c c }
& \multicolumn{3}{c}{\textsc{ice}-India} & \textsc{coraal} & \textsc{fwp}\\ 
\cmidrule(lr){2-4}\cmidrule(lr){5-5}\cmidrule(lr){6-6}
Approach & ROC-AUC & AP & Prec@100 & Prec@100 & Prec@100\\
\hline\hline
\textsc{AutoGen} & 68.94 & 12.63 & 16.93 & - & -\\
\textsc{AutoID} & 74.90 & 15.24 & 17.87 & - & -\\
\textsc{ManualGen} & 86.83 & 25.77 & 31.63 & 57.88 & 58.71\\
\textsc{AutoID + ManualGen} & 76.34 & 19.95 & 24.30 & - & -\\
\textsc{CGEdit} & 84.92 & 27.48 & 32.50 & \textbf{67.41} & 68.00\\
\textsc{ManualGen + CGEdit} & \textbf{88.76} & \textbf{29.32} & \textbf{35.67} & 64.94 & \textbf{74.35}\\
\end{tabularx}
\caption{Area under precision-recall curve (ROC-AUC), average precision (AP), and precision@100 in percentages for feature detection on all three corpora. Results are averages over all features (10 in \textsc{ice}-India, 17 in \textsc{coraal} and \textsc{fwp}). Reported scores for \textsc{ice}-India are averaged from three runs with different random seeds. Best scores are bolded.}
\label{table:2}
\end{table*}

\textbf{Overall results.}
Table 1 presents performance of the proposed approach against baselines and prior work. \textsc{AutoGen} and \textsc{AutoID} perform the worst across metrics. \textsc{CGEdit} outperforms \textsc{ManualGen}, the best prior work on this task, by up to 10 points in Prec@100 scores for both AAE datasets, \textsc{coraal} and \textsc{fwp}. Combining the training sets of \textsc{ManualGen} and \textsc{CGEdit} yielded the best performance, consistently outperforming \textsc{ManualGen} by about 4 points across metrics in \textsc{ICE-India} and by about 10-15 points in Prec@100 scores for both \textsc{coraal} and \textsc{fwp}. These gains can't simply be attributed to more training data, as combining \textsc{AutoID} and \textsc{ManualGen} training sets did not improve performance.

Better performance on AAE corpora may be due to a few variables: a higher number of AAE features means a larger total training set; larger AAE corpora mean more target corpus n-grams; the selected AAE features may be easier to distinguish or more prevalent than the IndE ones. Discrepancies between \textsc{coraal} and \textsc{fwp} are likely due to different feature prevalences. 

\textbf{Results by feature.}
Feature difficulty is similar across approaches; invariant features are easier to detect (i.e. focus \emph{itself} in IndE; \emph{finna} in AAE), while features with long-distance dependencies are more difficult (i.e. double object construction in AAE). See Appendix \ref{sec:app5} for complete results.

\section{Replicating Prior Sociolinguistic Work}
We recreate three recent studies of \textsc{coraal} where original authors manually annotated AAE morphosyntactic features and analyzed correlations between feature frequency and speaker metadata (i.e. gender, region, socioeconomic status). 
We used the combined \textsc{ManualGen + CGEdit} model and Classify \& Count (CC, summing hard classifications \cite{bella10}) to calculate per-speaker feature frequency.\footnote{In early experiments we tested the
\citet{saerens02} EM algorithm and PCC \citep{bella10} to improve frequency estimation, but found few improvements.} The same subsets of features and \textsc{coraal} were used as in previous work when possible; detailed results are in Appendix \ref{sec:app5}.

\begin{figure}[t]
\centering
\fbox{\includegraphics[width=.45\textwidth]{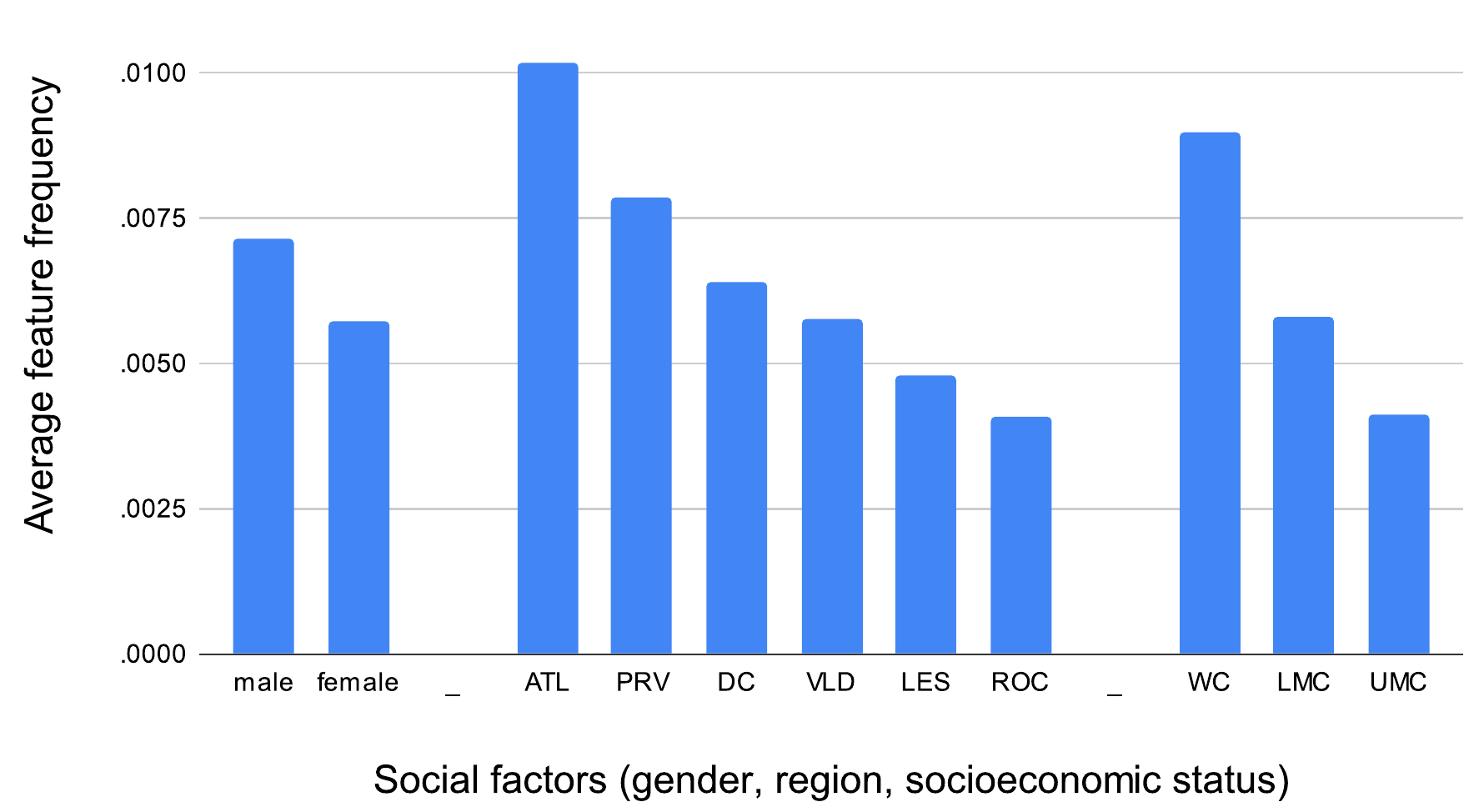}}
\caption{African American English feature variation by speaker's social factor, across all of \textsc{coraal}. Regions are Atlanta, GA; Princeville, NC; Washington, DC; Valdosta, GA; Lower East Side, NY; Rochester, NY; socioeconomic classes are Working Class, Lower Middle Class, Upper Middle Class.}
\label{fig:chart1}
\end{figure}

\citet{koenecke20} annotated 35 morphosyntactic features in 150 utterances. We confirm their conclusions that average feature frequency was lowest in Rochester, followed by DC, then Princeville; and lower among male speakers than female. 

\citet{cukor-avila19} looked at 3 features over 14,506 utterances. They qualitatively found considerable variation in feature use between speakers, even when within the same age group. We confirm this quantitatively: standard deviation between speakers within an age group is larger than standard deviation between age group means.%\bto{the last sentence is ambiguous for what exactly is having its SD taken. some readings of this are mathematically impossible via Law of Total Variation. do you mean the SD among age group means?}

\citet{grieser19} examined 14 features over 18,553 utterances. We confirm findings that age and socioeconomic status are negatively correlated with feature use. \citeauthor{grieser19} also found that being male was weakly correlated with feature use; interestingly, our results agree when we look at all 17 features or all of \textsc{coraal}, but not when we look at the same feature and data subsets as Grieser. This may indicate how small sample size (in terms of both features and datasets) can skew results. 

See Figure \ref{fig:chart1} for average frequencies of our 17 features in all 152,069 utterances of \textsc{coraal}, broken down by several social factors of the speaker. Feature detection at this scale is only possible with automatic methods, and allows researchers to draw more reliable conclusions about language use.

\section{Discussion and Future Work}
We propose a corpus-driven and manually-filtered approach to generate contrast sets for morphosyntactic feature detection in low-resource language varieties,
which may be useful for novel sociolinguistic analysis in future work.
This approach may be extendable to datasets with other nonstandard language varieties
(e.g.\ \textsc{ICE} with 14 English varieties \citep{greenbaum96},
\textsc{QADI} with 18 Arabic varieties \citep{abdelali21}, 
Corpus del Español with 21 Spanish varieties \citep{davies2016},
or Masakhane's African language collection, currently under development \citep{nekoto20}),
in addition to social media corpora, which are largely unlabeled and could benefit from automatic methods. 

Additionally, while we only examined automatic identification of noisy negatives, 
future work might explore automatic identification of reliable negatives by using an apt word representation and distance function to obtain unlabeled examples which are least similar to the positives \citep{bekker20}. Other extensions might consider adding manual filtering to an automatic identification approach, such as filtering through and identifying the nearest unlabeled examples that are true negatives, instead of identifying reliable (e.g. distant) negatives.

\section{Ethical Considerations and Broader Impact}
Our objective is to expand the linguistic coverage of NLP tools to include marginalized language varieties, so that they may also benefit from the linguistic analysis made possible by methodological innovation. We hope to aid both sociolinguistic and corpus linguistic researchers studying nonstandard language use. 

Since language varieties, including the ones examined in this study, may correlate with the national origin or ethnicity of the speaker and linguistic feature frequency may correlate with social factors, such as gender or socioeconomic status, there is a risk of automatic feature detection being used to infer personal information about a speaker \citep{kroger22, chancellor19, veronese19}. Our study has sought to show that there is a correlation between language use and social factors, but does not support any claims about the accuracy or ethics of using linguistic feature frequency to predict a given social factor. 

There is not a one-to-one mapping of feature frequency to ethnicity, socioeconomic status, or any other social factor. Two speakers with the same set of social factors may exhibit different feature frequencies; life circumstances do not deterministically produce linguistic competence. In addition, linguistic competence does not deterministically produce feature frequency. Every speaker has the ability to style-shift and thus use linguistic features to varying degrees for a given context, exhibiting a range of feature frequencies throughout their spoken interactions \citep{sharma17, sharma18}. There are many factors that may influence observed feature frequency, including pragmatic context, register, topic, relationship between the speakers, relationship to one's own identity, and so on. This complex relationship between language production and external factors should be considered when using this technology. 

\section*{Acknowledgements}
We would like to thank Devyani Sharma for help with accessing the \textsc{ice}-India corpus and Claudia Lange's annotations, with permission. We would also like to thank the UMass AAE group as well as anonymous reviewers from the First Field Matters Workshop for their helpful comments and feedback.
This work was supported by National Science Foundation grants 1845576 and 2042939;
any opinions, findings, and conclusions or recommendations expressed in this material are those of the authors and do not necessarily reflect the views of the National Science Foundation.

\clearpage

\bibliography{anthology,custom}
\bibliographystyle{acl_natbib}

\clearpage
\appendix

\section{Feature inventories}
\label{sec:app1}

\begin{table}[h!]
\small
\centering
\begin{tabularx}{\textwidth}{l l l}
\hline
\textbf{Level}&\textbf{IndE Feature} & \textbf{Example utterance}\\
\hline
Noun phrase & Non-initial existential \emph{there} & library facility was not \underline{there}\\
&Focus \emph{itself}&We are feeling tired now \underline{itself}\\
&Focus \emph{only}&I like dressing up I told you at the beginning \underline{only}\\
Verb phrase & Zero copula & Everybody (is) so worried about the exams\\
Sentence level & Left dislocation&\underline{we elders}, we don't have much time to converse\\
&Resumptive subject pronoun&the father, sometimes \underline{he} is unemployed\\
&Resumptive object pronoun&also pickles, we eat \underline{it} with this jaggery and lot of butter\\
&Topicalized object (argument) & \underline{brothers and sisters} you have\\
&Topicalized non-argument constituent&\underline{with your child} you have come\\
&Invariant tag \emph{no/na/isn't it} & both works same hours, \underline{isn't it}?\\
\hline
\end{tabularx}
\caption{Features of Indian English used in our study.}
\end{table}

\begin{table}[h!]
\small
\centering
\begin{tabularx}{\textwidth}{l l l l}
\hline
\textbf{Level}&\textbf{Grammatical domain} &\textbf{AAE Feature} &\textbf{Example utterance}\\
\hline
Noun phrase & Pronominal case & Zero possessive \emph{-'s} & go over my grandmama('s) house \\
Verb phrase& Copula deletion & Zero copula & she (is) the folk around here\\
& Tense marking & Double marked/overregularized & she \underline{likeded} me the best\\
& Aspect marking & Habitual \emph{be}&I just \underline{be} liking the beat\\
& & Resultant \emph{done}&you \underline{done} lost your mind\\
& Other verbal markers& \emph{finna}&she's \underline{finna} have a baby\\
& & \emph{come}&she \underline{come} grabbing me on my shirt\\
& & Double modal & he \underline{might could} really get our minds\\
& Negation & Negative concord & I \underline{ain't} doing \underline{nothing} wrong\\
& & Negative auxiliary inversion & \underline{don't nobody} know what I had\\
& & Non-inverted negative concord & \underline{nobody don't} say nothing\\
& & Preverbal negator \emph{ain't} & I \underline{ain't} doing nothing wrong\\
Sentence level & Subject-verb agreement & Zero 3rd p sg present tense \emph{-s} & I don't know if it count(s)\\
& & \emph{is/was}-generalization & they \underline{is} die hard Laker fans\\
& Number marking & Zero plural \emph{-s} & about four or five month(s)\\
& Ditransitive constructions & Double-object construction & I got \underline{me} my own car\\
& Interrogative constructions & \emph{Wh}-question & \underline{what} they was doing?\\
\hline
\end{tabularx}
\caption{Features of African American English used in our study.}
\end{table}

\clearpage
\section{Approach descriptions}
\label{sec:app2}

\subsection{Proposed approach}
A positive example $p$ is defined as $(x_1, x_2, ..., x_n)$ where $x_i$ is a subtoken. For each positive example $p$:
\begin{enumerate}
    \item A 3-gram instance $t$ in $p$ is defined as $(x_i, x_{i+1}, x_{i+2})$. For each 3-gram instance $t$ in $p$:
    \begin{enumerate}
        \item For each $n \in \{2, 3, 4\}$, find the 3 most frequent n-grams from the corpus where, for each n-gram $t'$, the set difference between $set(t)$ and $set(t')$ is at most one subtoken.
        \item Create perturbed examples by swapping $t$ for $t'$. These perturbed examples may or may not have the feature.
    \end{enumerate}
    \item Randomly order the perturbed examples.
    \item Manually filter and label the perturbed examples; examples that pass the filter should not have invalid subtoken combinations, positive examples should unambiguously have the feature, and negative examples should unambiguously not have the feature. Examples that pass the filter (positive or negative) may be ungrammatical. Stop after 2 positives and 3 negatives have passed the filter. Including the original positive example $p$, you should have 3 positives and 3 negatives. 
\end{enumerate}

We provide here an example of our approach. For the feature zero copula, we are given $p =$ He on the five dollar. We generate:
\begin{table}[h!]
\small
\begin{tabularx}{.45\textwidth}{l}
\hline
\textbf{Perturbed example}\\
\hline
He on the last five\\
He on the five\\
on the other five dollar\\
He on the five hundred dollar\\
He was on the dollar\\
on the five dollar\\
the on five dollar\\
He and five on the dollar\\
He was on the five dollar\\
He on the five dollar bill\\
He beating on the five dollar\\
He on the dollar\\
He on the other dollar\\
He on five dollar\\
He the five dollar\\
He on five dollar bill\\
was on the five dollar\\
\hline
\end{tabularx}
\end{table}

\newpage
The manually filtered contrast set looks like:
\begin{table}[h!]
\small
\begin{tabularx}{.45\textwidth}{l l}
\hline
\textbf{Example} & \textbf{Label}\\
\hline
He on the five dollar&1\\
He on the last five&1\\
He on the five&1\\
on the other five dollar&0\\
He was on the dollar&0\\
on the five dollar&0\\
\hline
\end{tabularx}
\end{table}

\subsection{Manual generation}
Given a positive example $p$, manually construct a negative example by modifying $p$ so they are (1) semantically-similar MAE versions, and (2) do not have the feature.

\subsection{Automatic generation}
For each positive example $p$:
\begin{enumerate}
    \item Randomly choose n-gram order, where n is some value 0 < n < length($p$) - 1.
    \item Split positive example into sequential non-overlapping n-grams from left to right. If length of sentence isn’t a multiple of n, then the remaining words form an additional m-gram (m < n).
    \item Randomly shuffle the list of n-grams.
    \item Repeat steps 1-3 until you have three distinct shuffled negative examples per positive example.\footnote{Number of negatives per positive was a tuned hyperparameter.}
\end{enumerate}

\subsection{Automatic identification}
Randomly choose unlabeled examples from target corpus and label them as the negative examples. Five negatives are chosen per positive example.\footnote{Number of negatives per positive was a tuned hyperparameter.}

%\clearpage
\section{Extended results and figures}
\label{sec:app5}
Tables 4, 5, and 6 are per-feature results for Indian English features in \textsc{ice}-India. Tables 7 and 8 are per-feature results for African American English features in \textsc{coraal} and \textsc{fwp}. Tables 9, 10, and 11 are standard deviation scores for Indian English features in \textsc{ice}-India. Figures 4, 5, and 6 are detailed results from replicating prior sociolinguistic work. 

\begin{table*}[!htbp]
\centering
\small
\begin{tabularx}{.8\textwidth}{l c c c c c c}
& \multicolumn{6}{c}{ROC-AUC}\\
\cmidrule(lr){2-7}
Feature & \textsc{AutoG.} & \textsc{AutoID} & \textsc{MnlG.} & \makecell{\textsc{AutoID}\\\textsc{+MnlG.}} & \textsc{CGEdit} & \makecell{\textsc{MnlG.}\\\textsc{+CGEdit}} \\
\hline\hline
Non-init. exist. \emph{there}& 91.14 & 90.47 & 89.88 & 89.74 & 95.46 & 89.03\\
Focus \emph{itself}& 94.08 & 98.02 & 98.70 & 97.58 & 99.49 & 99.89 \\
Focus \emph{only} & 85.38 & 97.00 & 98.94 & 95.40 & 96.72 & 99.02 \\
Zero copula & 53.28 & 61.82 & 73.75 & 67.77 & 73.79 & 75.61 \\
Left dislocation& 64.17 & 70.18 & 93.13 & 69.32 & 89.92 & 93.14 \\
Res. subject pronoun & 72.81 & 70.03 & 93.60 & 67.92 & 88.32 & 89.94 \\
Res. object pronoun & 67.49 & 70.46 & 86.87 & 78.24 & 86.44 & 88.93 \\
Topic. object (arg.) & 63.20 & 59.17 & 76.72 & 54.28 & 72.08 & 81.30 \\
Topic. non-arg. const.& 44.90 & 55.48 & 69.24 & 55.55 & 59.99 & 79.54 \\
Invar. tag \emph{no/na/isn't it} & 52.96 & 76.37 & 87.46 & 87.55 & 86.95 & 91.24\\
\textbf{Macro average} & 68.94 & 74.90 & 86.83 & 76.34 & 84.92 & 88.76\\
\cmidrule(lr){1-7}
\end{tabularx}
\caption{ROC-AUC results on \textsc{ice}-India, averaged over 3 runs.}
\end{table*}

\begin{table*}[t]
\centering
\small
\begin{tabularx}{.8\textwidth}{l c c c c c c}
& \multicolumn{6}{c}{AP}\\
\cmidrule(lr){2-7}
Feature & \textsc{AutoG.} & \textsc{AutoID} & \textsc{MnlG.} & \makecell{\textsc{AutoID}\\\textsc{+MnlG.}} & \textsc{CGEdit} & \makecell{\textsc{MnlG.}\\\textsc{+CGEdit}} \\
\hline\hline
Non-init. exist. \emph{there}& 46.56 & 41.32 & 53.16 & 51.84 & 61.11 & 59.56\\
Focus \emph{itself}& 39.99 & 40.16 & 74.76 & 72.76 & 78.12 & 75.14 \\
Focus \emph{only} & 24.23 & 32.74 & 40.04 & 28.12 & 41.10 & 44.31 \\
Zero copula & 01.78 & 04.96 & 02.05 & 04.19 & 03.88 & 02.95 \\
Left dislocation& 02.78 & 05.70 & 25.78 & 09.47 & 23.07 & 26.63 \\
Res. subject pronoun & 03.68 & 03.57 & 21.72 & 07.55 & 20.64 & 20.50 \\
Res. object pronoun & 00.24 & 01.58 & 02.47 & 00.93 & 02.96 & 05.66 \\
Topic. object (arg.) & 02.04 & 15.95 & 06.99 & 02.13 & 06.00 & 10.16 \\
Topic. non-arg. const.& 01.11 & 02.53 & 03.78 & 02.26 & 02.65 & 06.10 \\
Invar. tag \emph{no/na/isn't it} & 03.89 & 04.96 & 26.95 & 20.26 & 37.26 & 42.18\\
\textbf{Macro average} & 12.63 & 15.24 & 25.77 & 19.95 & 27.48 & 29.32\\
\cmidrule(lr){1-7}
\end{tabularx}
\caption{AP results on \textsc{ice}-India, averaged over 3 runs.}
\end{table*}

\begin{table*}[t]
\centering
\small
\begin{tabularx}{.8\textwidth}{l c c c c c c}
& \multicolumn{6}{c}{Prec@100}\\
\cmidrule(lr){2-7}
Feature & \textsc{AutoG.} & \textsc{AutoID} & \textsc{MnlG.} & \makecell{\textsc{AutoID}\\\textsc{+MnlG.}} & \textsc{CGEdit} & \makecell{\textsc{MnlG.}\\\textsc{+CGEdit}} \\
\hline\hline
Non-init. exist. \emph{there}& 78.33 & 74.00 & 86.00 & 82.00 & 84.33 & 87.00\\
Focus \emph{itself}& 15.67 & 18.67 & 28.00 & 25.00 & 28.00 & 28.00 \\
Focus \emph{only} & 34.33 & 41.33 & 48.33 & 39.67 & 45.00 & 48.33 \\
Zero copula & 03.33 & 01.67 & 03.33 & 05.00 & 03.00 & 05.33 \\
Left dislocation& 08.33 & 18.33 & 46.33 & 27.00 & 42.67 & 42.00 \\
Res. subject pronoun & 09.67 & 13.67 & 39.00 & 24.67 & 36.00 & 31.67 \\
Res. object pronoun & 00.00 & 01.00 & 03.67 & 01.67 & 04.67 & 08.33 \\
Topic. object (arg.) & 05.67 & 03.00 & 15.00 & 06.67 & 12.33& 19.33 \\
Topic. non-arg. const.& 01.33 & 01.00 & 07.33 & 06.33 & 07.00 & 13.67 \\
Invar. tag \emph{no/na/isn't it} & 12.67 & 06.00 & 39.33 & 25.00 & 62.00 & 73.00\\
\textbf{Macro average} & 16.93 & 17.87 & 31.63 & 24.30 & 32.50 & 35.67\\
\cmidrule(lr){1-7}
\end{tabularx}
\caption{Prec@100 results on \textsc{ice}-India, averaged over 3 runs.Prec@100 results on \textsc{coraal}.
Note that if there are less than 100 instances of a certain feature (e.g. \textit{finna} occurs only 35 times in this dataset, confirmed via keyword search), then its Prec@100 score will have an upper bound of less than 1.}
\end{table*}

\begin{table*}[t]
\centering
\small
\begin{tabularx}{.55\textwidth}{l c c c }
& \multicolumn{3}{c}{Prec@100}\\
\cmidrule(lr){2-4}
Feature & \textsc{MnlG.} & \textsc{CGEdit} & \makecell{\textsc{MnlG.}\\\textsc{+CGEdit}} \\
\hline\hline
Zero possessive \emph{-'s}& 030.0 & 071.0 & 088.0 \\
Zero copula& 089.0 & 100. 0& 100.0 \\
Double marked & 024.0 & 031.0 & 045.0  \\
Habitual \emph{be} & 100.0 & 100.0 & 100.0 \\
Resultant \emph{done}& 089.0 & 097.0 & 097.0 \\
\emph{finna} & 035.0 & 035.0 & 035.0\\
\emph{come} & 011.0 & 016.0 & 015.0 \\
Double modal & 014.0 & 014.0 & 013.0\\
Negative concord& 100.0 & 096.0 & 077.0 \\
Neg. auxiliary inversion & 078.0 & 096.0 & 089.0\\
Non-inverted neg. concord & 009.0 & 010.0 & 012.0  \\
Preverbal negator \emph{ain't} & 100.0 & 100.0 & 100.0\\
Zero 3rd p sg pres. tense \emph{-s} & 096.0 & 100.0 & 098.0\\
\emph{is/was}-generalization & 063.0 & 100.0 & 100.0 \\
Zero plural \emph{-s} & 017.0 & 062.0 & 059.0\\
Double-object construction & 050.0 & 030.0 & 018.0  \\
\emph{Wh}-question & 079.0 & 088.0 & 058.0 \\
\textbf{Macro average} & 057.9 & 067.4 & 064.9 \\
\cmidrule(lr){1-4}
\end{tabularx}
\caption{Prec@100 results on \textsc{coraal}.
Note that if there are less than 100 instances of a certain feature (e.g. \textit{finna} occurs only 35 times in this dataset, confirmed via keyword search), then its Prec@100 score will have an upper bound of less than 1.}
\end{table*}

\begin{table*}[t]
\centering
\small
\begin{tabularx}{.55\textwidth}{l c c c }
& \multicolumn{3}{c}{Prec@100}\\
\cmidrule(lr){2-4}
Feature & \textsc{MnlG.} & \textsc{CGEdit} & \makecell{\textsc{MnlG.}\\\textsc{+CGEdit}} \\
\hline\hline
Zero possessive \emph{-'s}& 011.0 & 042.0 & 026.0 \\
Zero copula& 097.0 & 099.0 & 100.0 \\
Double marked & 053.0 & 049.0 & 095.0  \\
Habitual \emph{be} & 078.0 & 099.0 & 097.0 \\
Resultant \emph{done}& 093.0 & 100.0 & 100.0 \\
\emph{finna} & 000.0 & 000.0 & 000.0\\
\emph{come} & 001.0 & 050.0 & 082.0 \\
Double modal & 004.0 & 005.0 & 004.0\\
Negative concord& 100.0 & 100.0 & 100.0 \\
Neg. auxiliary inversion & 093.0 & 100.0 & 100.0\\
Non-inverted neg. concord & 015.0 & 024.0 & 056.0  \\
Preverbal negator \emph{ain't} & 100.0 & 100.0 & 100.0\\
Zero 3rd p sg pres. tense \emph{-s} & 100.0 & 100.0 & 100.0\\
\emph{is/was}-generalization & 100.0 & 100.0 & 100.0  \\
Zero plural \emph{-s} & 024.0 & 070.0 & 096.0\\
Double-object construction & 036.0 & 028.0 & 020.0  \\
\emph{Wh}-question & 093.0 & 090.0 & 088.0 \\
\textbf{Macro average} & 058.7 & 068.0 & 074.4 \\
\cmidrule(lr){1-4}
\end{tabularx}
\caption{Prec@100 results on \textsc{fwp}.
Note that if there are less than 100 instances of a certain feature (e.g. \textit{finna} occurs 0 times in this dataset, confirmed via keyword search), then its Prec@100 score will have an upper bound of less than 1.}
\end{table*}

\begin{table*}[t]
\centering
\small
\begin{tabularx}{.8\textwidth}{l c c c c c c}
& \multicolumn{6}{c}{ROC-AUC Standard Deviation}\\
\cmidrule(lr){2-7}
Feature & \textsc{AutoG.} & \textsc{AutoID} & \textsc{MnlG.} & \makecell{\textsc{AutoID}\\\textsc{+MnlG.}} & \textsc{CGEdit} & \makecell{\textsc{MnlG.}\\\textsc{+CGEdit}} \\
\hline\hline
Non-init. exist. \emph{there}& 03.29 & 00.69 & 00.65 & 07.39 & 01.89 & 08.42\\
Focus \emph{itself}& 03.38 & 00.54 & 00.42 & 00.45 & 00.47 & 00.03 \\
Focus \emph{only} & 06.40 & 01.59 & 00.66 & 01.25 & 02.74 & 00.48 \\
Zero copula & 04.63 & 03.80 & 07.95 & 04.71 & 06.87 & 01.04 \\
Left dislocation& 07.90 & 01.83 & 01.24 & 16.00 & 01.62 & 00.78 \\
Res. subject pronoun & 04.62 & 07.10 & 00.39 & 17.13 & 04.77 & 05.24 \\
Res. object pronoun & 04.73 & 06.15 & 05.66 & 07.79 & 01.77 & 00.70 \\
Topic. object (arg.) & 06.20 & 02.88& 10.93 & 06.49 & 04.89 & 05.39 \\
Topic. non-arg. const.& 03.25 & 05.52 & 03.87 & 01.79 & 05.57 & 03.31 \\
Invar. tag \emph{no/na/isn't it} & 07.64 & 04.35 & 03.04 & 01.59 & 10.77 & 04.97\\
\textbf{Macro average} & 05.20 & 03.45 & 03.48 & 06.46 & 04.14 & 03.04\\
\cmidrule(lr){1-7}
\end{tabularx}
\caption{Standard deviation of ROC-AUC results on \textsc{ice}-India over 3 runs.}
\end{table*}

\begin{table*}[t]
\centering
\small
\begin{tabularx}{.8\textwidth}{l c c c c c c}
& \multicolumn{6}{c}{AP Standard Deviation}\\
\cmidrule(lr){2-7}
Feature & \textsc{AutoG.} & \textsc{AutoID} & \textsc{MnlG.} & \makecell{\textsc{AutoID}\\\textsc{+MnlG.}} & \textsc{CGEdit} & \makecell{\textsc{MnlG.}\\\textsc{+CGEdit}} \\
\hline\hline
Non-init. exist. \emph{there}& 09.52 & 03.07 & 04.32 & 15.13 &09.13 & 08.09\\
Focus \emph{itself}& 09.87 & 11.30 & 03.44 & 08.26 & 04.30 & 08.19 \\
Focus \emph{only} & 08.36 & 02.62 & 05.68 & 08.01 & 04.74 & 00.43\\
Zero copula & 01.79 & 05.45 & 01.22 & 02.07 & 01.50  & 01.36 \\
Left dislocation& 00.80 & 01.31 & 04.90 & 05.84 & 01.36  & 00.78 \\
Res. subject pronoun & 00.70 & 03.12 & 07.30 & 05.54 &  08.82  & 04.91 \\
Res. object pronoun & 00.07 & 01.77 & 00.72 & 00.89 & 00.65 & 01.83 \\
Topic. object (arg.) & 01.29 & 25.05& 02.46 & 00.57  & 01.31 & 01.18 \\
Topic. non-arg. const.& 00.13 & 01.93 & 00.99 & 00.96  & 00.93 & 00.39 \\
Invar. tag \emph{no/na/isn't it} & 00.73 & 03.02 & 13.96  & 07.02 & 25.90 & 16.98 \\
\textbf{Macro average} & 03.33 & 05.86 & 04.50  & 05.43 & 05.86 & 04.41\\
\cmidrule(lr){1-7}
\end{tabularx}
\caption{Standard deviation of AP results on \textsc{ice}-India over 3 runs.}
\end{table*}

\begin{table*}[t]
\centering
\small
\begin{tabularx}{.8\textwidth}{l c c c c c c}
& \multicolumn{6}{c}{Prec@100 Standard Deviation}\\
\cmidrule(lr){2-7}
Feature & \textsc{AutoG.} & \textsc{AutoID} & \textsc{MnlG.} & \makecell{\textsc{AutoID}\\\textsc{+MnlG.}} & \textsc{CGEdit} & \makecell{\textsc{MnlG.}\\\textsc{+CGEdit}} \\
\hline\hline
Non-init. exist. \emph{there}& 08.02 & 07.00 & 04.00 & 12.90 & 04.16 & 03.61\\
Focus \emph{itself}& 03.51 & 04.04 & 00.00 & 31.19 & 00.00 & 00.00 \\
Focus \emph{only} & 06.03 & 04.16 & 06.43 & 07.13 & 05.57 & 05.51 \\
Zero copula & 01.15 & 01.53 & 02.08 & 01.30 & 03.00 &01.53 \\
Left dislocation& 04.04 & 06.66 & 05.20 & 34.27 & 05.51 & 02.65 \\
Res. subject pronoun & 04.51 & 03.21 & 14.73 & 21.81 & 17.69 & 07.09 \\
Res. object pronoun & 00.00 & 00.00 & 01.15 & 02.89 & 00.58 & 02.52 \\
Topic. object (arg.) & 03.79 & 02.65& 05.20 & 06.48 & 02.31 & 03.51 \\
Topic. non-arg. const.& 00.58 & 00.00 & 03.21 & 07.57 & 03.00 & 03.79 \\
Invar. tag \emph{no/na/isn't it} & 04.16 & 04.36 & 16.20 & 38.91 & 25.51 & 17.09\\
\textbf{Macro average} & 03.58 & 03.36 & 06.15 & 16.45 & 06.73 & 04.73\\
\cmidrule(lr){1-7}
\end{tabularx}
\caption{Standard deviation of Prec@100 results on \textsc{ice}-India over 3 runs.}
\end{table*}

\clearpage

\begin{figure*}[!h]
\centering
\fbox{\includegraphics[scale=.5]{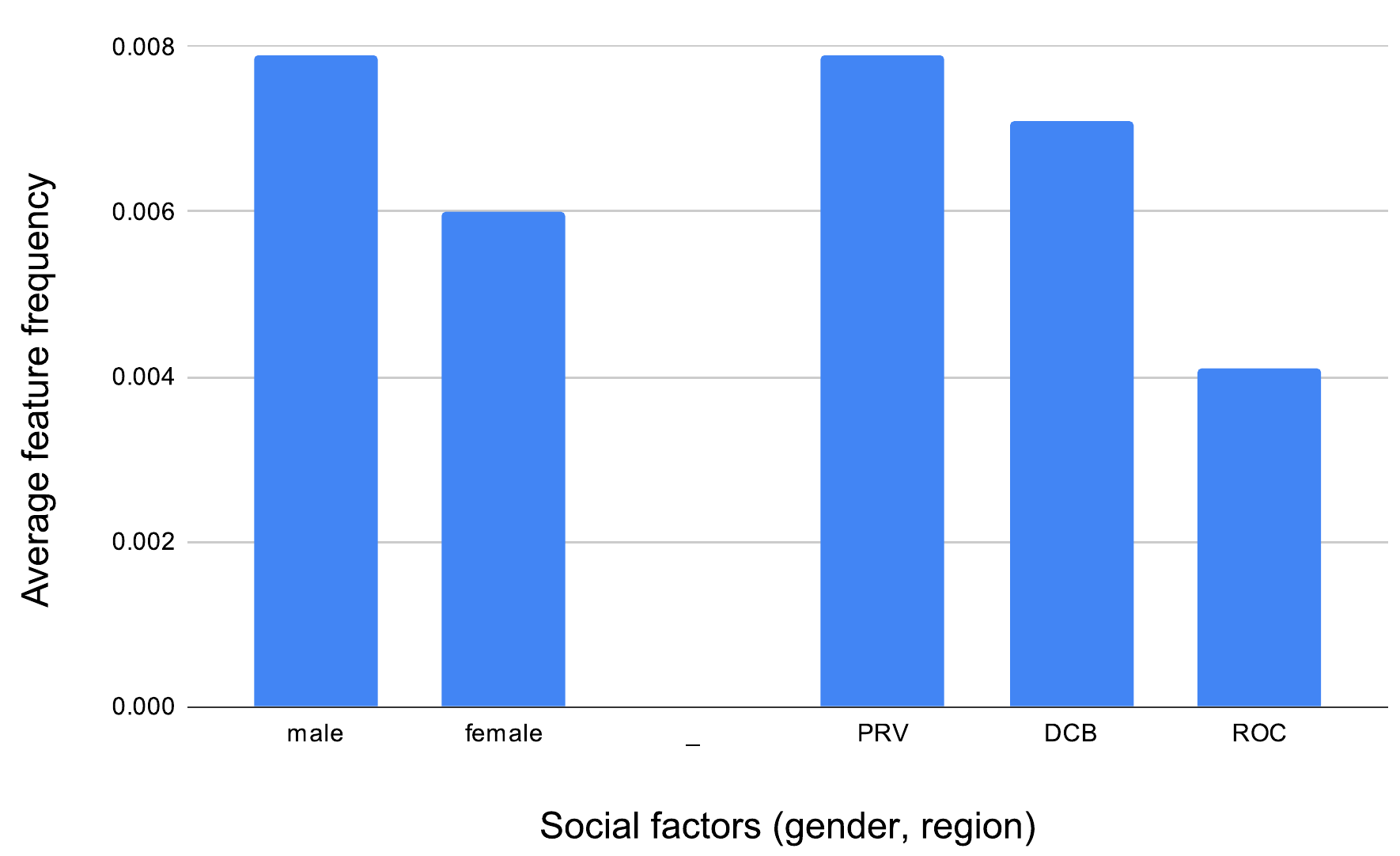}}
\caption{Confirming results from \citet{koenecke20}. Examined 17 features over entire DCB, PRV, and ROC subcorpora. We find higher feature frequencies among male speakers than female speakers; and highest feature frequency in Princeville, followed by DC, and then Rochester.}
\end{figure*}

\begin{figure*}[h!]
\centering
\fbox{\includegraphics[scale=.5]{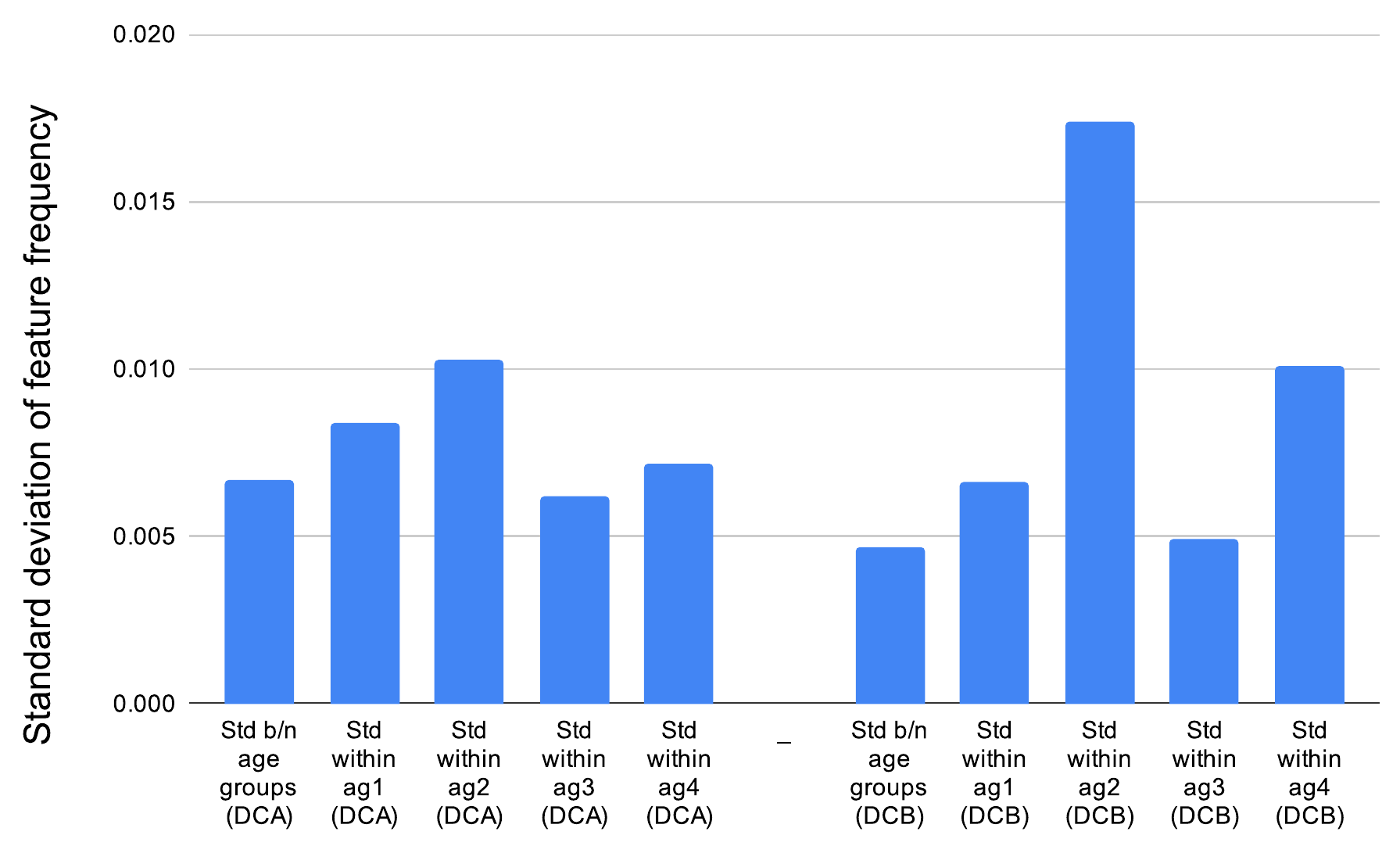}}
\caption{Confirming results from \citet{cukor-avila19}. Examined 3 features over files specified in their study from DCA and DCB subcorpora. Ag1 corresponds to ages less than 20, ag2 corresponds to ages 20-29, ag3 corresponds to 30-50, and ag4 corresponds to 50+. We find that standard deviation between speakers in an age group is equal to or larger than standard deviation between age groups.}
\end{figure*}

\clearpage
\begin{figure*}[h!]
\centering
\fbox{\includegraphics[scale=.5]{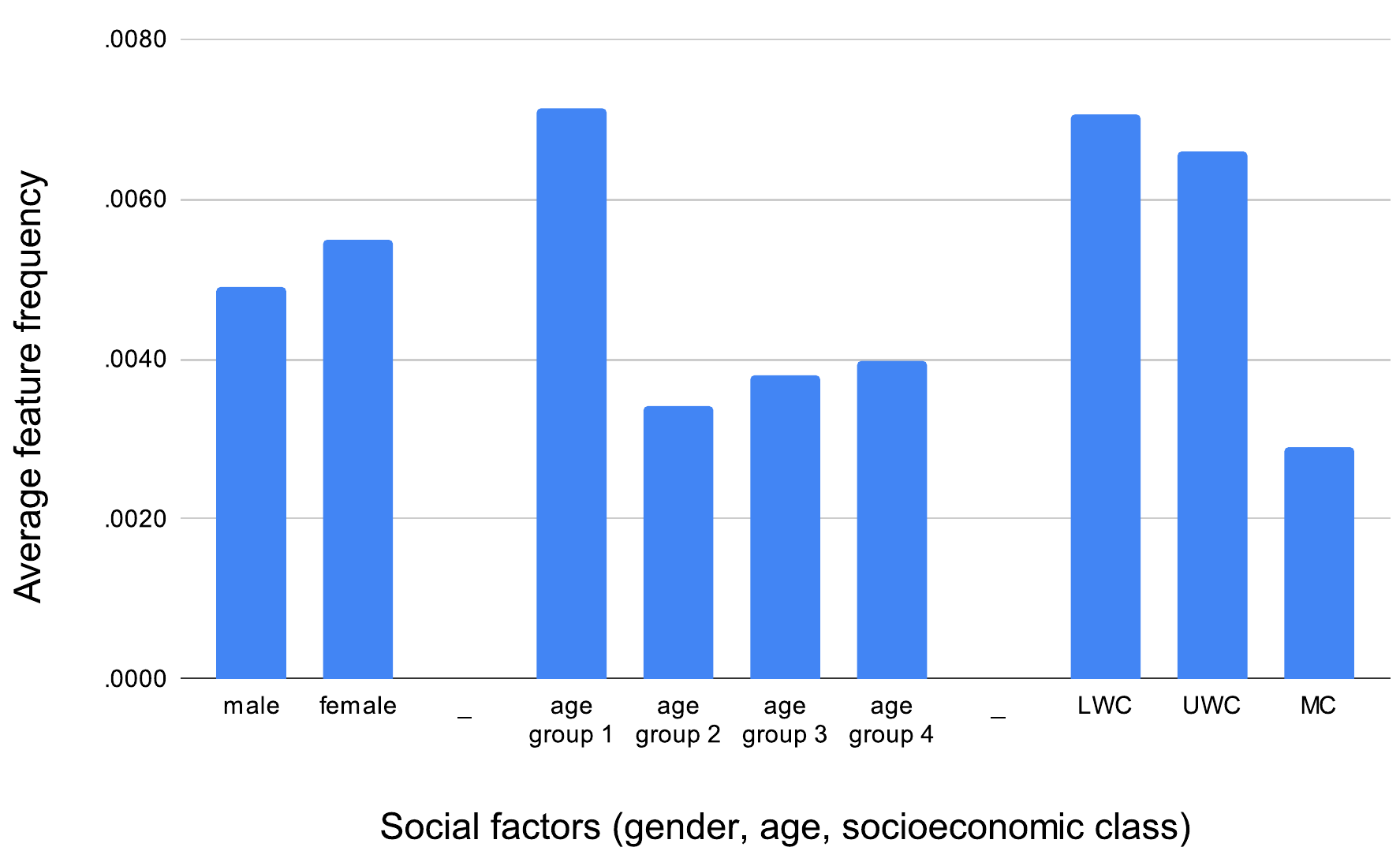}}
\caption{Confirming results from \citet{grieser19}. Examined 14 features over files specified in their study from DCA subcorpus. Age group 1 corresponds to ages less than 20, age group 2 corresponds to ages 20-29, age group 3 corresponds to 30-50, and age group 4 corresponds to 50+; the socioeconomic classes, from left to right, are Lower Working Class, Upper Working Class, and Middle Class. We find that age and socioeconomic status are negatively correlated with feature use. We find that men have a slightly lower average feature frequency; however, when looking at all of \textsc{coraal} for all of our features, we confirm that men have a higher average feature frequency. This is perhaps an example of how small sample size can skew results.}
\end{figure*}

\end{document}